\documentclass[10pt,twocolumn,letterpaper]{article}

\usepackage{wacv}              

\usepackage{graphicx}
\usepackage{algorithm}
\usepackage{algpseudocode}
\usepackage{amsmath}
\usepackage{amssymb}
\usepackage{booktabs}
\usepackage{diagbox}
\usepackage{bm}
\usepackage[accsupp]{axessibility}

\usepackage[pagebackref,breaklinks,colorlinks]{hyperref}

\usepackage[capitalize]{cleveref}
\crefname{section}{Sec.}{Secs.}
\Crefname{section}{Section}{Sections}
\Crefname{table}{Table}{Tables}
\crefname{table}{Tab.}{Tabs.}

\def\wacvPaperID{623} 
\def\confName{WACV}
\def\confYear{2025}

\begin{document}

\title{Active Learning with Context Sampling and One-vs-Rest Entropy for Semantic Segmentation}

\author{
    Fei Wu$^{1}$, Pablo Marquez-Neila$^{1}$, Hedyeh Rafii-Tari$^{2}$, Raphael Sznitman$^{1}$\\
    $^{1}$University of Bern \\ 
    $^{2}$Johnson \& Johnson \\
    {\tt\small \{fei.wu,pablo.marquez,raphael.sznitman\}@unibe.ch}\\
    {\tt\small hrafiita@its.jnj.com}
}
\maketitle

\newif\ifdraft
 \drafttrue

\definecolor{orange}{rgb}{1,0.5,0}
\definecolor{darkgreen}{RGB}{0,100,0} 

\ifdraft
 \newcommand{\RS}[1]{{\color{red}{\bf RS: #1}}}
 \newcommand{\rs}[1]{{\color{red}#1}}
 \newcommand{\PMN}[1]{{\color{orange}{\bf PMN: #1}}}
 \newcommand{\pmn}[1]{{\color{orange}#1}}
 \newcommand{\fei}[1]{{\color{blue}{\bf fei: #1}}}
 \else
 \newcommand{\RS}[1]{{\color{red}{}}}
 \newcommand{\rs}[1]{#1}
 \newcommand{\PMN}[1]{{\color{red}{}}}
 \newcommand{\pmn}[1]{#1}
 \newcommand{\fei}[1]{{\color{blue}{\bf fei: #1}}}
\fi

\newcommand{\real}{\mathbb{R}}
\newcommand{\x}{\mathbf{x}}
\newcommand{\X}{\mathbf{X}}
\newcommand{\Y}{\mathbf{Y}}
\newcommand{\s}{\mathbf{s}}
\newcommand{\SP}{\mathbf{S}}
\newcommand{\z}{\mathbf{z}}
\newcommand{\y}{\mathbf{y}}
\renewcommand{\a}{\mathbf{a}}
\renewcommand{\k}{\mathbf{k}}
\newcommand{\haty}{\hat{\y}}
\newcommand{\w}{\mathbf{w}}
\renewcommand{\d}{\mathbf{d}}
\newcommand{\A}{\mathcal{A}}
\newcommand{\U}{\mathcal{U}}
\newcommand{\Q}{\mathcal{Q}}
\newcommand{\cX}{\mathcal{X}}
\newcommand{\cS}{\mathcal{S}}
\newcommand{\cY}{\mathcal{Y}}
\newcommand{\cZ}{\mathcal{Z}}
\newcommand{\cP}{\mathcal{P}}
\newcommand{\C}{\mathcal{C}}

\algnewcommand{\LeftComment}[1]{\(\triangleright\) #1}
\makeatletter
\algnewcommand{\LineComment}[1]{\Statex \hskip\ALG@thistlm \(\triangleright\) #1}
\algnewcommand{\IndentLineComment}[1]{\Statex \hskip\ALG@tlm \(\triangleright\) #1}
\makeatother
\algrenewcommand\algorithmicrequire{\textbf{Input:}}
\algrenewcommand\algorithmicensure{\textbf{Output:}}

\newcommand{\argtopk}[1]{\mathop{\operatorname{arg\,top-#1}}}

\newcommand{\RSamp}{{\emph{Random}}}
\newcommand{\BvSBSamp}{{\emph{BvSB}}}
\newcommand{\RevisitingSPSamp}{{\emph{Revisiting SP}}}
\newcommand{\CBALSamp}{{\emph{CBAL}}}
\newcommand{\PixelBalSamp}{{\emph{PixelBal}}}
\newcommand{\OurSamp}{{\emph{OREAL}}}
\newcommand{\Ours}{{\bfemph}}

\newcommand{\mycolor}{black} 
\newcommand{\NewText}[1]{\textcolor{\mycolor}{#1}}

\newcommand{\rebuttalcolor}{black} 
\newcommand{\RebuttalText}[1]{\textcolor{\rebuttalcolor}{#1}}

\newcommand{\MeanAgg}{{$mean_{agg}$}}
\newcommand{\MaxAgg}{{$max_{agg}$}}

\newcommand{\TX}[1]{\underline{#1}}
\newcommand{\TY}[1]{\textbf{#1}}
\begin{abstract}
Multi-class semantic segmentation remains a cornerstone challenge in computer vision. Yet, dataset creation remains excessively demanding in time and effort, especially for specialized domains. Active Learning (AL) mitigates this challenge by selecting data points for annotation strategically. However, existing patch-based AL methods often overlook boundary pixels' critical information, essential for accurate segmentation. We present \OurSamp{}, a novel patch-based AL method designed for multi-class semantic segmentation. \OurSamp{} enhances boundary detection by employing maximum aggregation of pixel-wise uncertainty scores. Additionally, we introduce \emph{one-vs-rest entropy}, a novel uncertainty score function that computes class-wise uncertainties while achieving implicit class balancing during dataset creation. Comprehensive experiments across diverse datasets and model architectures validate our hypothesis. 

\end{abstract}

\section{Introduction}

Multi-class image segmentation is one of the main problems in computer vision, with significant implications for a wide range of applications, including medical imaging, autonomous driving, and environmental monitoring. However, annotating segmentation maps in images is a time-consuming and costly task, especially for applications that require domain-specific knowledge. In an effort to alleviate this annotation burden, active learning (AL) has been proposed due to its objective of training the best possible machine learning model using the least amount of annotations collected iteratively. With its potential benefits, AL methods for image segmentation have been developed over the last few years~\cite{BRATS_AL,DSAL,suggestive_ann,synthetic_image_AL,vae_AL,revisiting,RIPU,viewAL,self_consistency,RL_AL,MetaBox,MEAL,DIAL,CEREALS,adaptiveSP, multi_label_AL, COWAL, region_AL}.

\begin{figure}[h]
    \includegraphics[width=\linewidth]{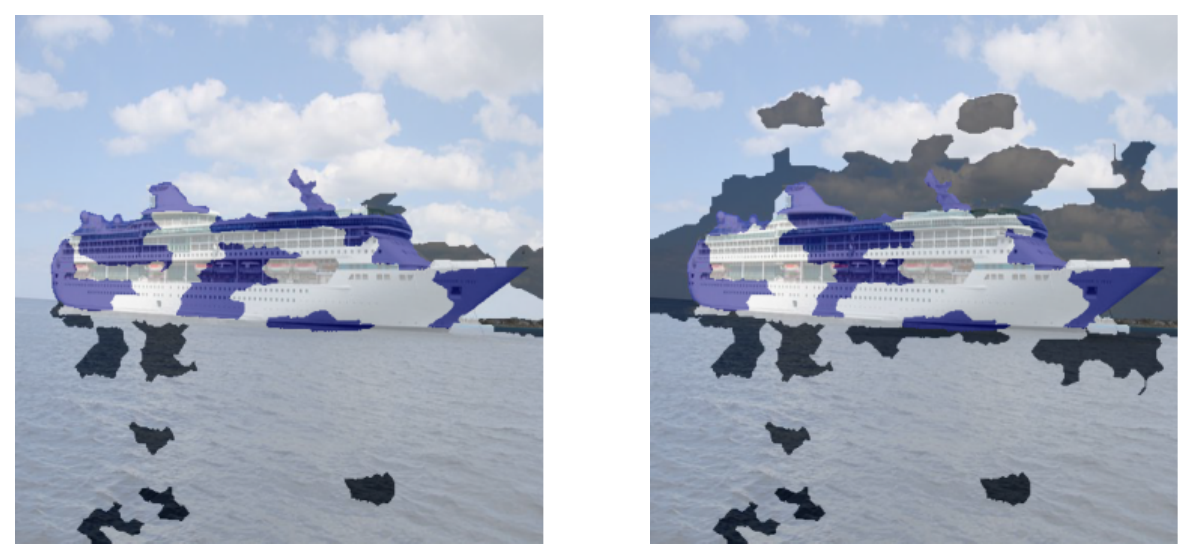}
   \caption{Context sampling with patch-based active learning for semantic segmentation. Mean aggregation (left) ignores patches at the boundaries of the object of interest, while maximum aggregation (right) encourages the sampling of these boundaries. This simple modification of the aggregation function resulted in an improvement of 3~percentage points in mIoU.}
\label{fig: intro_example}
\end{figure}

Broadly, AL methods for image segmentation can be categorized into two: those that sample images~\cite{BRATS_AL, suggestive_ann, DSAL, synthetic_image_AL, vae_AL, COWAL} and those that sample image patches \cite{revisiting, RIPU, viewAL, MetaBox, CEREALS, adaptiveSP, multi_label_AL, MEAL, RL_AL, self_consistency, DIAL, region_AL}. Image-level sampling methods identify which images from an unlabeled set are most beneficial to annotate, then requiring a human annotator to provide labels for each pixel of the selected images. On the other hand, patch sampling methods identify which patches (\ie,~groups of pixels) are more beneficial to annotate across images in the unlabeled dataset, and require a human annotator to provide labels for the pixels of the selected patches. In general, patch-based methods have been shown to perform better, largely due to reduced annotation times~\cite{CEREALS}.



\NewText{Patch-sampling methods are more efficient than image sampling in part because they eliminate the need to annotate large amounts of uninformative background pixels. However, not all background pixels are uninformative. Empirical evidence suggests that the background pixels that are in close proximity to the object of interest are particularly beneficial in enhancing the quality of the predicted segmentations, as object boundaries are often the most challenging areas. Interestingly, existing patch-based methods do not prioritize the sampling of the background patches that correspond to these informative boundary pixels.}

\NewText{
In order to make patch-based methods aware of boundary areas, a simple but crucial observation is noting that boundary pixels often exhibit extreme statistics that differ significantly from those observed in non-boundary pixels. For example, the magnitude of the image gradient has been widely used in classical computer vision for boundary detection, as it is locally maximized in boundary pixels. Similarly, uncertainty-based scores, commonly used in AL, often reach their maximum value in boundary pixels, as these areas tend to be the most challenging. However, when scoring patches for AL, existing patch-based methods simply average the scores of the individual pixels within each patch. This averaging process diminishes the influence of boundary pixels, rendering the patch scores more uniform and thus less effective at identifying patches with informative pixels. We claim that extreme values, and not average values, are critical to properly describe the informativeness of patches. By just substituting the mean aggregation with a maximum aggregation of pixel-wise uncertainty scores, we achieved a better sampling of boundary patches (Fig.~\ref{fig: intro_example}), which in turn resulted in improved segmentation quality. Therefore, we propose using aggregation functions that are sensitive to extreme values inside each patch as a simple but effective manner of improving patch-based AL methods.}

While maximum aggregation improves boundary sampling, selecting patches with a simple uncertainty score function (\eg,~entropy) does not guarantee optimal results in the context of AL for multi-class semantic segmentation. Besides maximizing uncertainty, a second important criterion is class balancing of the labeled set, as it is well-known that an imbalanced labeled set can have a significant negative impact in a multi-class setting. This has led numerous works to try to address this challenge by combining model predictions with uncertainty measures~\cite{active_class_IL, AL_imbalanced_dataset, algo_selection_imbalanced, CBAL, clinical_imbalance, imbalanced_cold_start, minority_classes,revisiting, RIPU,multi_label_AL}. 
Instead, we introduce a novel uncertainty score function, namely \emph{one-vs-rest entropy}, that leverages class-wise uncertainties to implicitly incorporate class-balancing.

\NewText{
Finally, we propose a new patch-based AL method \OurSamp{} for multi-class semantic segmentation. \OurSamp{} combines maximum aggregation for improved boundary sampling and the \emph{one-vs-rest entropy} uncertainty score for implicit class balancing. We support our claims with an extensive collection of experiments using four datasets, three architectures (including a vision transformer), and two labeling schemes for superpixel patches.}



\section{Related Work}
We review the most relevant work in this area.

\subsection{Semantic Segmentation}
Semantic segmentation is a process that classifies each image pixel into a predefined image category. \RebuttalText{While several weak labeling methods exist for reducing annotation cost \cite{point_labels, building_point_labels, bbox_labels, weak_labels} on this task, we focus on using AL methods in this work.} A plethora of AL methods for semantic segmentation have been advanced, which predominantly bifurcate into two streams: image-level sampling \cite{BRATS_AL,DSAL,suggestive_ann,synthetic_image_AL,vae_AL,COWAL} and region-specific sampling \cite{revisiting,RIPU,viewAL,self_consistency,RL_AL,MetaBox,MEAL,DIAL,CEREALS,adaptiveSP,multi_label_AL}. Research delineated in \cite{CEREALS} underscores the preeminence of employing patch-based sampling rather than whole-image sampling, with an observation that reduced patch sizes correlate with enhanced results. Although there is a subset of work that engages with non-rectangular, superpixel-based patches, its advantage over rectangular patches remains debated~\cite{region_AL}.

\subsection{Superpixel Sampling}

Various patch shapes are possible in patch-based methods, including rectangular patches~\cite{self_consistency, RL_AL, MetaBox, MEAL, CEREALS, DIAL, RIPU} and superpixels~\cite{revisiting, region_AL, viewAL, adaptiveSP, multi_label_AL}. Unlike rectangular patches, superpixels adhere to the object boundaries and group pixels that are \RebuttalText{visually} similar~\cite{SEEDS}, implicitly assuming that all pixels within a superpixel share a unique label. This property greatly simplifies the annotation process of superpixels~\cite{revisiting}, as the user only needs to provide a single label for each superpixel, which is then assigned to all of its pixels.
This labeling method is known as the dominant label. 
Considering that the annotation cost is given by the number of user clicks,~\cite{revisiting}~showed that superpixels achieved significant improvements compared to rectangular patches at a reduced annotation cost. Further developments in superpixel patch sampling, as outlined in subsequent research~\cite{adaptiveSP}, focused on merging superpixels with similar predictions to expand the coverage of the single label scheme per superpixel. A recent work~\cite{multi_label_AL} proposed a weak labeling scheme for superpixels instead of employing a single dominant label. The weak labeling scheme requires the annotator to identify all present classes within the patch without specifying the corresponding pixels. Although the central component of~\cite{multi_label_AL} is the weak labeling scheme, their proposed sampling strategy also works with the dominant labeling scheme.\RebuttalText{~\cite{region_AL} applies a Conditional Random Field (CRF) on top of the model prediction and simply selects superpixels based on the entropy obtained from the CRF output. 
~\cite{viewAL} proposes a viewpoint entropy that exploits the unique property of viewpoint consistency in multi-view datasets and is thus not applicable to standard image datasets.}
Although our method OREAL is adaptable to a variety of patch shapes, we chose to utilize superpixel patches with dominant label annotation in our experiments due to their simplicity and proven advantages in reducing annotation costs and enhancing result quality.


\subsection{Class Imbalanced Datasets}
Class imbalance is a prevalent issue in AL, and numerous methods have been developed to balance the labeled image pool, with the majority of them designed for classification tasks \cite{active_class_IL, AL_imbalanced_dataset, algo_selection_imbalanced, CBAL, clinical_imbalance, imbalanced_cold_start, minority_classes}. Conversely, only a few methods target segmentation tasks~\cite{revisiting, RIPU,multi_label_AL}.
To achieve an equal number of images per class in a dataset, it is necessary to identify unlabeled samples likely to belong to the underrepresented or tail classes. Methods that attempt to do so typically rely on the model prediction \cite{RIPU,revisiting,CBAL,multi_label_AL} or an image embedding \cite{active_class_IL,AL_imbalanced_dataset,clinical_imbalance,minority_classes} to determine the image class. 

\RebuttalText{Methods that rely on image embedding~\cite{active_class_IL,AL_imbalanced_dataset,clinical_imbalance,minority_classes} compare the similarity of unlabeled images with labeled images from tail classes and choose the unlabeled images most similar to labeled tail class images.}
However, the majority of these methods evaluate the entire image for this purpose and are thus not comparable to the presented work, which focuses on superpixels as the annotation unit. \RebuttalText{On another hand, ~\cite{RIPU} prioritizes image regions with the highest number of predicted classes. Since, we chose to annotate each region with a single label, regions with a great number of different classes will suffer from this annotation scheme. We hence discard ~\cite{RIPU} in our comparison.}


\NewText{The novel AL method we introduce is thus compared} with \cite{CBAL}, \cite{revisiting}, and \cite{multi_label_AL}. \cite{CBAL} utilizes image entropy \cite{entropy} and the softmax output of an image to decide which image to sample. They create a vector of the count of images per class that should be added to the current training set to balance the number of images in each class. After processing an unlabeled image, the distance between its softmax probability output with a vector of count is calculated to determine the probability that it represents a tail class. The resulting distance is subtracted from the image entropy, and the images with the highest subtracted entropy are sampled first. 
\RebuttalText{In the case of \cite{revisiting}, they calculate a weight per class based on the predicted labels of superpixels}. This weight prioritizes tail classes and scales the uncertainty scores. They proceed to sample the superpixel with the highest weighted uncertainty. Likewise, \cite{multi_label_AL} adopts the same methodology as \cite{revisiting}, but computes the weight per class utilizing pixel-predicted probability, as opposed to superpixel-predicted labels. \RebuttalText{We follow a similar class balancing strategy than \cite{CBAL} by defining a vector of image count per class. However, unlike \cite{CBAL}, we use this vector to distribute the annotation budget per class. Then for each class, we rely on our introduced \emph{one-vs-rest entropy} to select the most uncertain samples per class. Unlike previous methods \cite{revisiting, multi_label_AL, CBAL} which combined the model entropy and the model prediction to select the most uncertain samples of tail classes, \emph{one-vs-rest entropy} is a standalone metric that finds the most uncertain samples w.r.t.~to a given class. We empirically show that this idea improves the model's ability to focus on underrepresented classes, leading to better overall performance in diverse tasks.}
\section{Method}

\begin{figure}[t]
    \includegraphics[width=\linewidth]{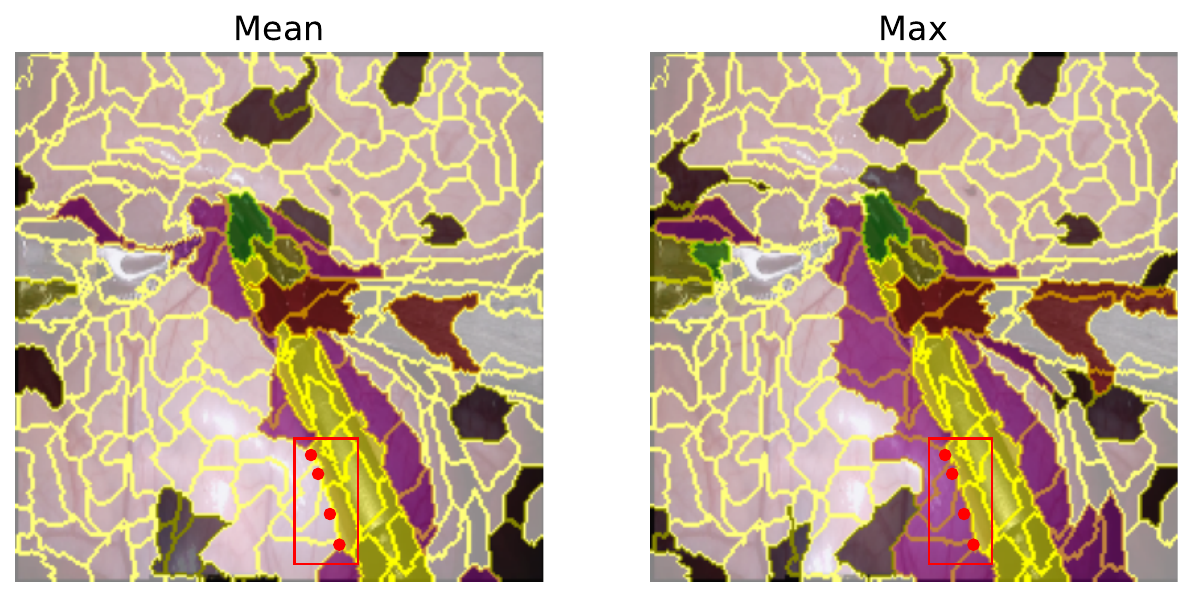}
   \caption{Comparison between selected superpixel regions (yellow borders) when the score of the superpixel is defined as the \emph{Mean} (left) or \emph{Max} (right) aggregation of its pixel score. The image is from the EndoVis \cite{endovis} dataset which displays a surgery tool.}
\label{fig: SP selection comparison}
\end{figure}

\label{sec: method}
A multi-class segmentation model is a function~$f:\cX\to\cY$ that takes an image $\x\in\cX$ and produces a label map $\y=f(\x)\in\cY$ where each pixel belongs to a set of $C$~classes,~$y_i\in\C = \{1,\ldots,C\}$. Training such a model requires pairs $(\x, \y')\in\cX\times\RebuttalText{\mathcal{Y'}}$ of images and their corresponding \RebuttalText{partially} annotated label maps. \RebuttalText{Unlike the predicted label maps~$\y$, the pixels of a partially annotated label map~$\y'\in\cY'$ belong to an extended set of classes $y_i\in\C'=\{\bot, 1, \ldots, C\}$ that include the special class~$\bot$ to indicate unlabelled pixels}.
Annotated label maps for training are expensive to obtain because annotation requires human intervention. To minimize the manual effort involved in annotating extensive image datasets, AL trains a model iteratively with existing labeled data and chooses new items from the pool of unlabeled data for annotation. In the context of patch-based AL for segmentation, the annotation item are image patches.

\subsection{Superpixels}
\label{sec: superpixel definition}
Following previous work~\cite{revisiting}, we choose superpixels over rectangular areas as our shape for image patches. The superpixels\RebuttalText{~$\{\mathcal{K}_i\}_{i=1}^{K}$} of an image~$\x$ form a disjoint partition of its pixels, where each superpixel\RebuttalText{~$\mathcal{K}_i\subset \mathbb{N}$} is a connected subset of the image pixels that exhibit high visual uniformity (Fig.~\ref{fig: SP selection comparison}). This uniformity increases the likelihood that all pixels within a superpixel belong to the same class. Consequently, a human annotator is only required to provide a single label for each superpixel, the most prevalent one. This approach reduces the annotation cost to one click per superpixel and has been demonstrated to yield superior results with a lower annotation cost than rectangular patches~\cite{revisiting}.

Patch-based AL requires an annotation strategy to select superpixels to annotate. Formally, at each AL iteration~$t$, the set of all available superpixels~$\cS$ in the training set is divided between the set of labeled superpixels\RebuttalText{~$\A_t \subset \cS\times \C$} and the set of unlabeled superpixels~$\U_t\subset \cS$. An \emph{annotation strategy function}~$\pi:2^\cS\times 2^{\cS\times\C}\to 2^\cS$ takes the unlabeled~$\U_t$ and labeled~$\A_t$ superpixels and suggests the set of superpixels~$\mathcal{Q}_t=\pi(\U_t, \A_t)\subset\U_t$ to annotate in the next iteration. After annotation, the superpixels in~$\mathcal{Q}_t$ and their labels are moved to~$\A_{t+1}$. The model is then trained with~$\A_{t+1}$, and the whole process is repeated. 

\subsection{Maximum Aggregation}
\label{sec: max agg}
\NewText{In practice, a score is assigned to each superpixel of $\U_t$, and $\pi$ selects the superpixels with the highest scores. Using the segmentation model~$f$, a score can be calculated for each pixel. For superpixel units, we define the score of a superpixel as the aggregation of its pixel scores. While the natural choice for this aggregation is the average of scores (\MeanAgg), we found out that using the maximum of pixel scores (\MaxAgg) to define the superpixel score has some interesting properties. Fig.~\ref{fig: SP selection comparison} shows two examples of selected superpixel regions when using the \MeanAgg~and \MaxAgg~approach. While \MeanAgg~and \MaxAgg~have both selected superpixels of the surgery tool, \MaxAgg~has further selected regions around it, creating context around the object. The reason \MaxAgg~can achieve this behavior is explained using the dots inside the red rectangle. These dots are pixels with high uncertainty located at the boundary of the surgery tool. They belong however to background superpixels. Using the \MeanAgg~to calculate the uncertainty score for these background regions will dilute the value of these pixels. However, \MaxAgg~will emphasize them, which results in these background regions being selected and thus creating a context around the object of interest.}

\subsection{One-vs-rest entropy}
\label{sec:class entropy}
\NewText{In addition, we also propose our own sampling strategy in the context of semantic segmentation}. Our proposed sampling strategy combines entropy maximization and class balancing as criteria for constructing the query set~$\mathcal{Q}$. Class balancing aims to keep the labels in~$\A_t$ as uniform as possible while entropy maximization aims to favor the most uncertain superpixels in general. We incorporate these two criteria into the \emph{one-vs-rest entropy} score by extending the uncertainty of a superpixel to a per-class level. If class~$c$ is underrepresented in~$\A_t$, the annotation strategy will favor superpixels most uncertain to belong to this class. However, uncertain samples for a class are not necessarily from the said class, possibly breaking the class balancing objective. We explain in Sec.~\ref{sec: ann_strat} how this balance is preserved.

Given the categorical distribution~$P_i$ for the pixel~$i$ predicted by the segmentation model~$f$, the \emph{one-vs-rest entropy} (or OVR entropy) for class~$c$ and pixel~$i$ is the entropy of a binarized representation of the distribution considering the class~$c$ vs.~the rest of the classes,
\begin{equation}
    H_c[i] = -P_i[c]\log P_i[c] - (1-P_i[c])\log \left(1-P_i[c]\right).
\end{equation}
Note that we omit the dependency on the image~$\x$ for clarity. 

The OVR entropy for a superpixel\RebuttalText{~$\mathcal{K}$} is computed using the maximum aggregation of its pixel OVR entropies. We therefore define the OVR entropy for class~$c$ and superpixel\RebuttalText{~$\mathcal{K}$} as
\begin{equation}
    \label{eq:superpixel_OVR_entropy}
    H_c[\mathcal{P}] = \max_{i\in\RebuttalText{\mathcal{K}}} H_c[i].
\end{equation}
A high OVR entropy indicates high uncertainty about this superpixel being from class~$c$, hence annotating this superpixel will improve the prediction accuracy for class~$c$.

\subsection{Class balancing}
\label{sec: annotation_strat}

Class balancing aims to maintain a uniform distribution of labels in the labeled set~$\A_t$ by favoring the selection of superpixels with a high probability of belonging to underrepresented classes. To keep class balance, our annotation strategy requires to have access to the \emph{class debt}, \ie,~the number of samples missing per class to reach class balance. To this end, our strategy computes a vector of required class counts~$\bm{\delta}$, where each element~$\delta_c$ indicates the number of superpixels of class~$c$ required to maximize class balancing. This involves maximizing

\begin{align}
\label{eq:optimize_delta}
\max_{\bm{\delta} \in \mathbb{N}^C} & \min_c n_c + \delta_c, \\
\nonumber
\textrm{s.t.} & \sum_c \delta_c = Q,
\end{align}

where $n_c$ is the number of superpixels in~$\A_t$ labeled with class~$c$, and $Q$~is the size of the query set~$\Q$. 

\subsection{Annotation strategy}
\label{sec: ann_strat}

Our annotation strategy first estimates the required number of superpixels per class with the class counts of Eq.~\eqref{eq:optimize_delta} and then selects the $\delta_c$~superpixels with the highest OVR entropy~$H_c$ for each class~$c$ iteratively. Alg.~\ref{alg: annotation strategy} details the procedure.

\begin{algorithm}
\caption{Annotation strategy $\pi$}
\label{alg: annotation strategy}
\begin{algorithmic}[1]
\Require $\U_t$ (set of unlabeled superpixels), $\A_t$ (set of labeled superpixels), $Q$ (number of superpixels to select for annotation)
\Ensure $\Q_t$ (selected superpixels for annotation)
\Function{SelectSuperpixels}{$\U_t$, $\A_t$, $Q$}
\State $\Q \leftarrow \emptyset$

\State Train the segmentation model~$f$ with data in $\A_t$
\State Apply $f$ over the images of $\U_t$ to compute pixel-wise class distributions
\State Compute $H_c[\cP] \quad \forall \cP\in\U_t$
\State $n_c\leftarrow \sum_{(\cP, y)\in\A_t} [y=c] \quad\forall c$
\Statex \Comment{{\footnotesize Class-wise counts of labels in~$\A_t$}}
\State $\bm{\delta} \leftarrow \Call{ItemsPerClass}{\mathbf{n}, Q}$ \Comment{{\footnotesize \NewText{See Appendix}}}
\For {$c\in\mathcal{C}$}
    \State $\Q \leftarrow \Q \cup {\displaystyle \argtopk{\delta_c}_{\cP\in\U_t}} \  H_c[\cP]$
    \Statex \Comment{{\footnotesize Pick the $\delta_c$~superpixels with the highest~$H_c$}}
\EndFor
\State \Return $\Q$
\EndFunction
\end{algorithmic}
\end{algorithm}

For high OVR entropy~$H_c$, the selected elements are the most uncertain for class~$c$. However, uncertain samples of class~$c$ do not guarantee that they are actually from that class, potentially resulting in a deviation from the predetermined~$\delta_c$. However, in practice, this discrepancy does not significantly impact class balance, as any deviation from~$\delta_c$ is compensated for in subsequent iterations. \NewText{See Appendix for more details}.
\begin{figure}[b!]
    \includegraphics[width=\linewidth]{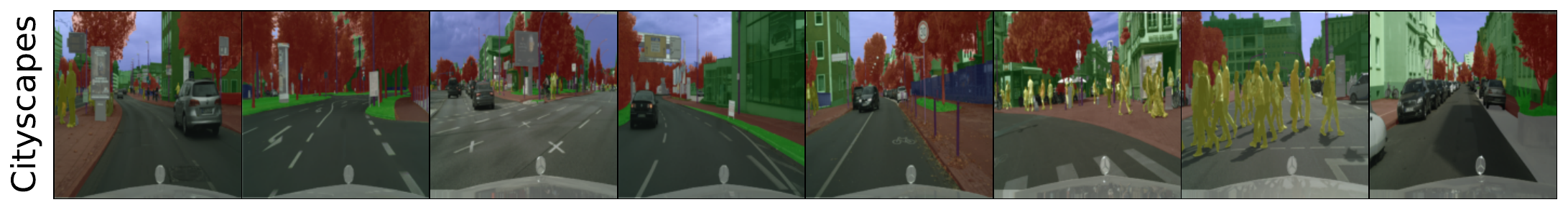}
    \includegraphics[width=\linewidth]{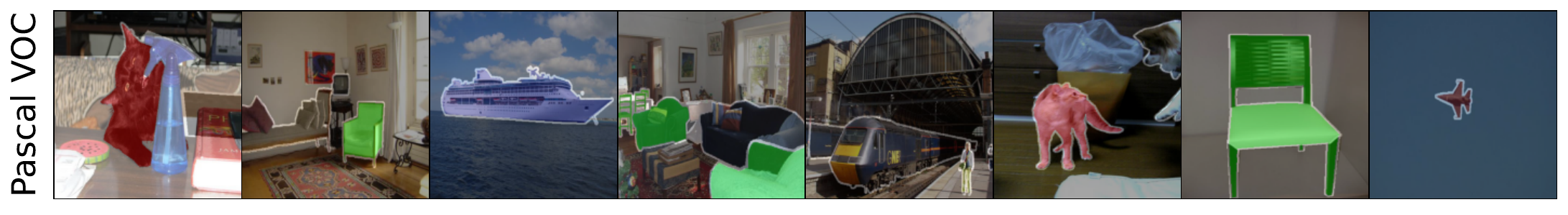}
    \medskip
    \includegraphics[width=\linewidth]{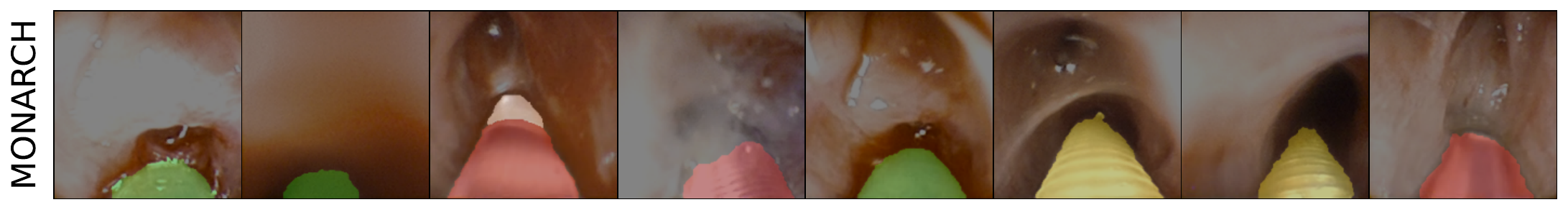}
    \includegraphics[width=\linewidth]{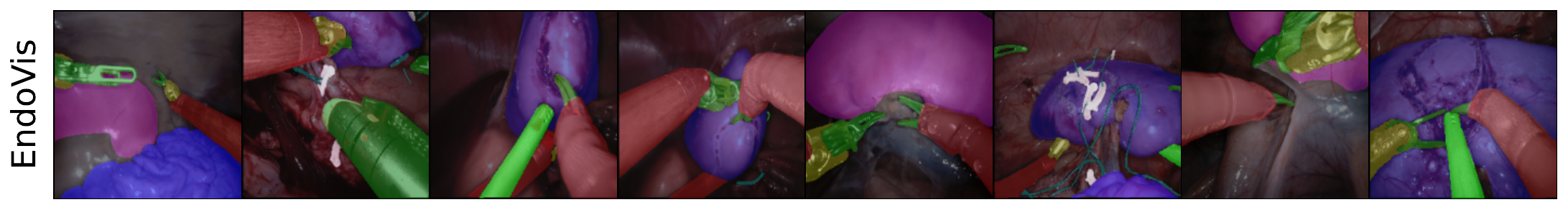}
   \caption{First row shows images from Cityscapes \cite{cityscapes} and Pascal VOC \cite{pascalVOC}, while second row shows images from MONARCH \cite{COWAL} and EndoVis \cite{endovis}. Images are overlayed by their segmentation mask with a color per class.}
\label{fig:auris}
\end{figure}

\section{Experiments}

In the following, we outline our experimental settings, implementation details, and evaluation protocol.

\subsection{Datasets \& Baselines} 
\label{sec: Dataset}

\begin{description}
    \item[Cityscapes~\cite{cityscapes}] Video sequences captured in urban environments across 50 cities. It comprises 19~categories, a training set of 2,975~images, a testing set of 1,525~images, and a validation set of 500~images. We retain the same training set and utilize the validation set as our test set.
    \item[Pascal VOC~\cite{pascalVOC}] Images from everyday scenes. It contains 20~foreground classes, with a training set comprising 1,464~images and a validation set consisting of 1,449~images. We retain the same training set and utilize the validation set as our test set.
    \item[MONARCH~\cite{COWAL}] Videos collected using the MONARCH{\texttrademark} platform, a robot-assisted bronchoscopy system intended for diagnostic and therapeutic procedures within the lung. It comprises  55~videos, 2,875~frames, and 7~foreground classes. The 55~videos have been divided into 33~videos comprising 1,727~frames for training and 22~videos comprising 1,148~frames for testing.
    \item[EndoVis~\cite{endovis}] Surgery videos featuring segmentation masks on surgical tools and human body organs. The dataset includes a total of 19~videos with 11~foreground classes, separated into 15~training videos consisting of 2,235~frames and 4~test videos consisting of 997~frames.  
\end{description}

Fig.~\ref{fig:auris} shows example images from each of the datasets used in this work. We compare our approach to a variety of both classical and state-of-the-art AL methods for segmentation such as \RSamp, \BvSBSamp~\cite{BvSB}, \RevisitingSPSamp~\cite{revisiting}, \PixelBalSamp~\cite{multi_label_AL}, and \CBALSamp~\cite{CBAL}.

\begin{figure*}[t]
    \includegraphics[width=0.5\linewidth]{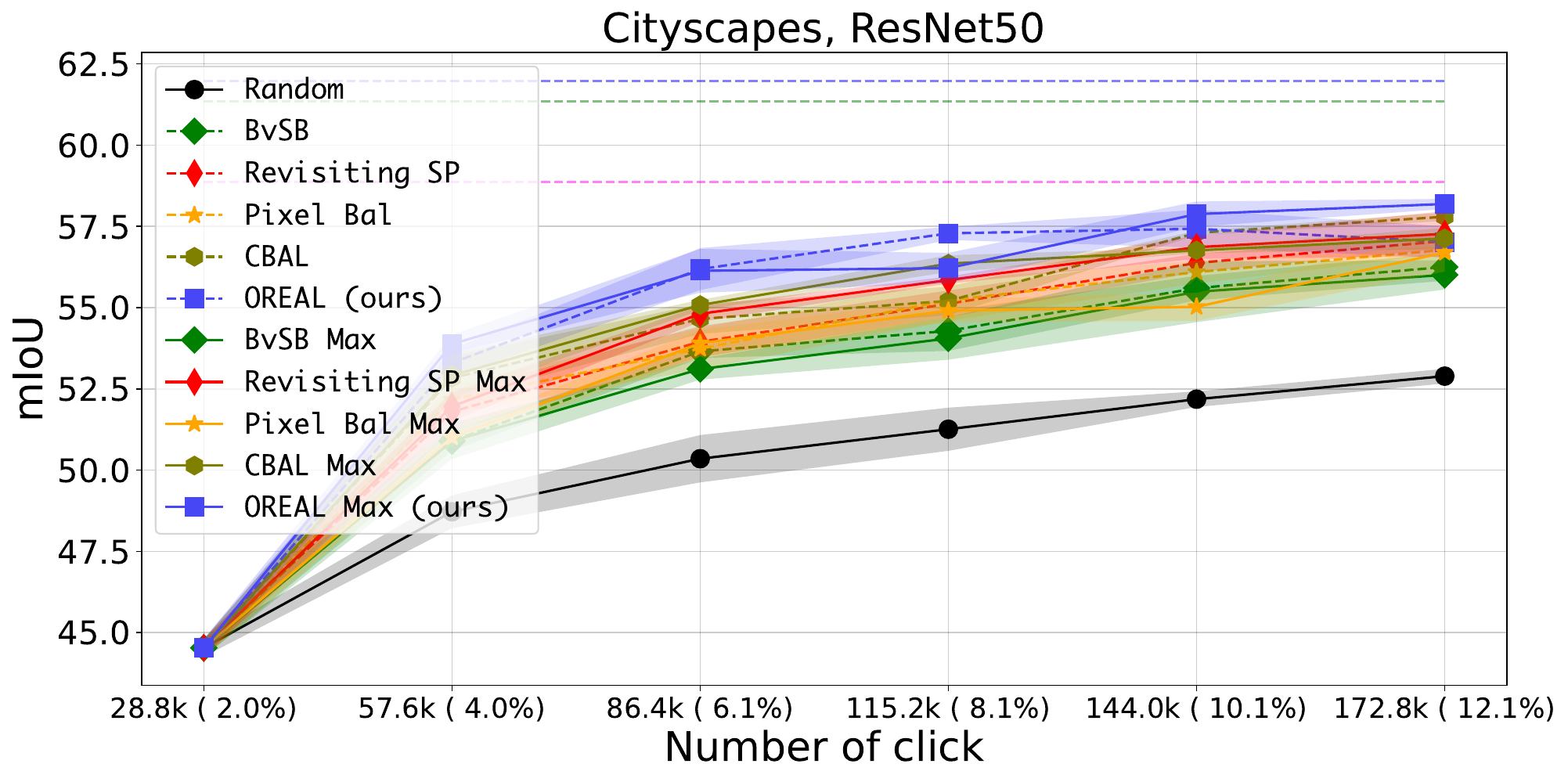}
    \includegraphics[width=0.5\linewidth]{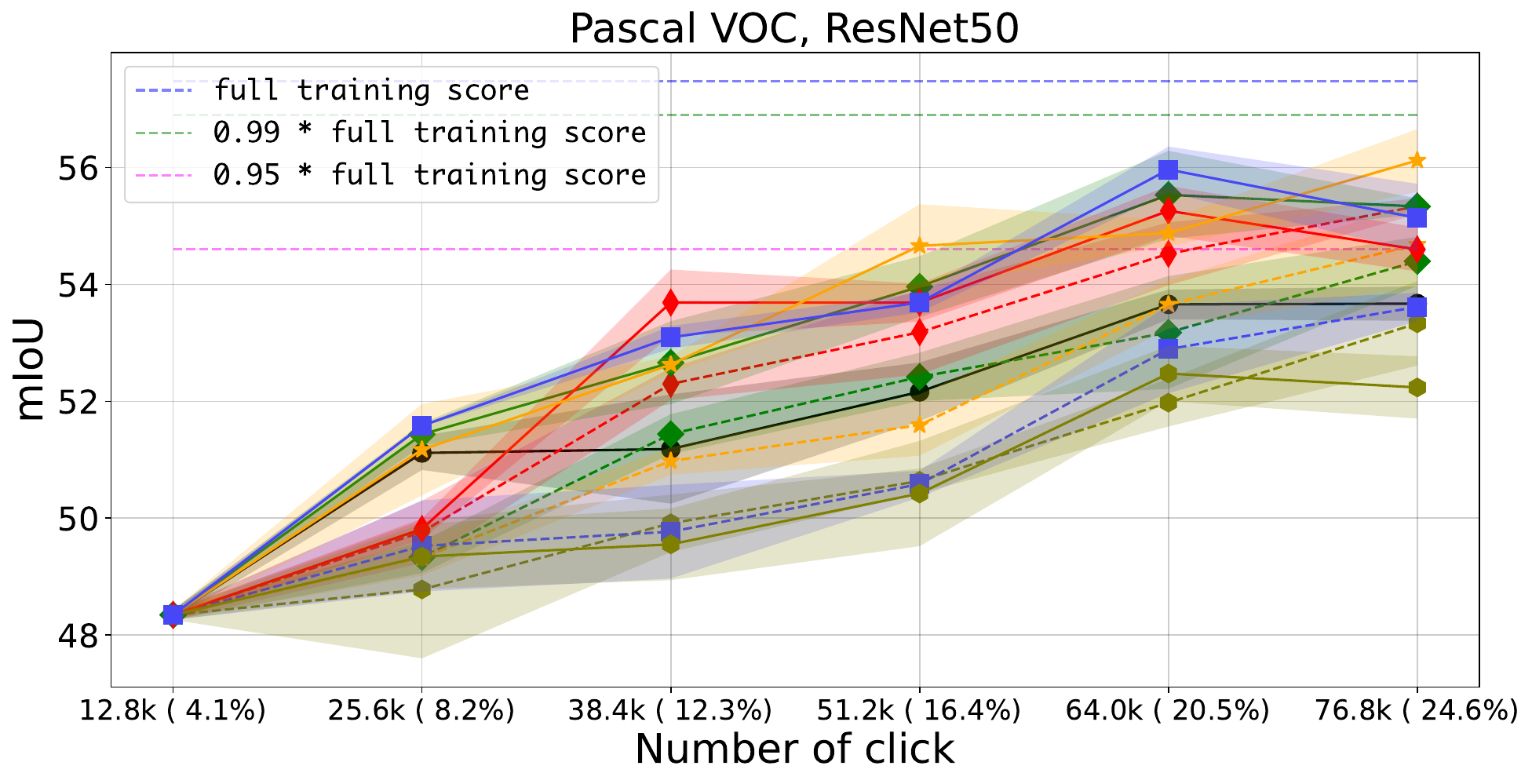}
    \medskip
    \includegraphics[width=0.5\linewidth]{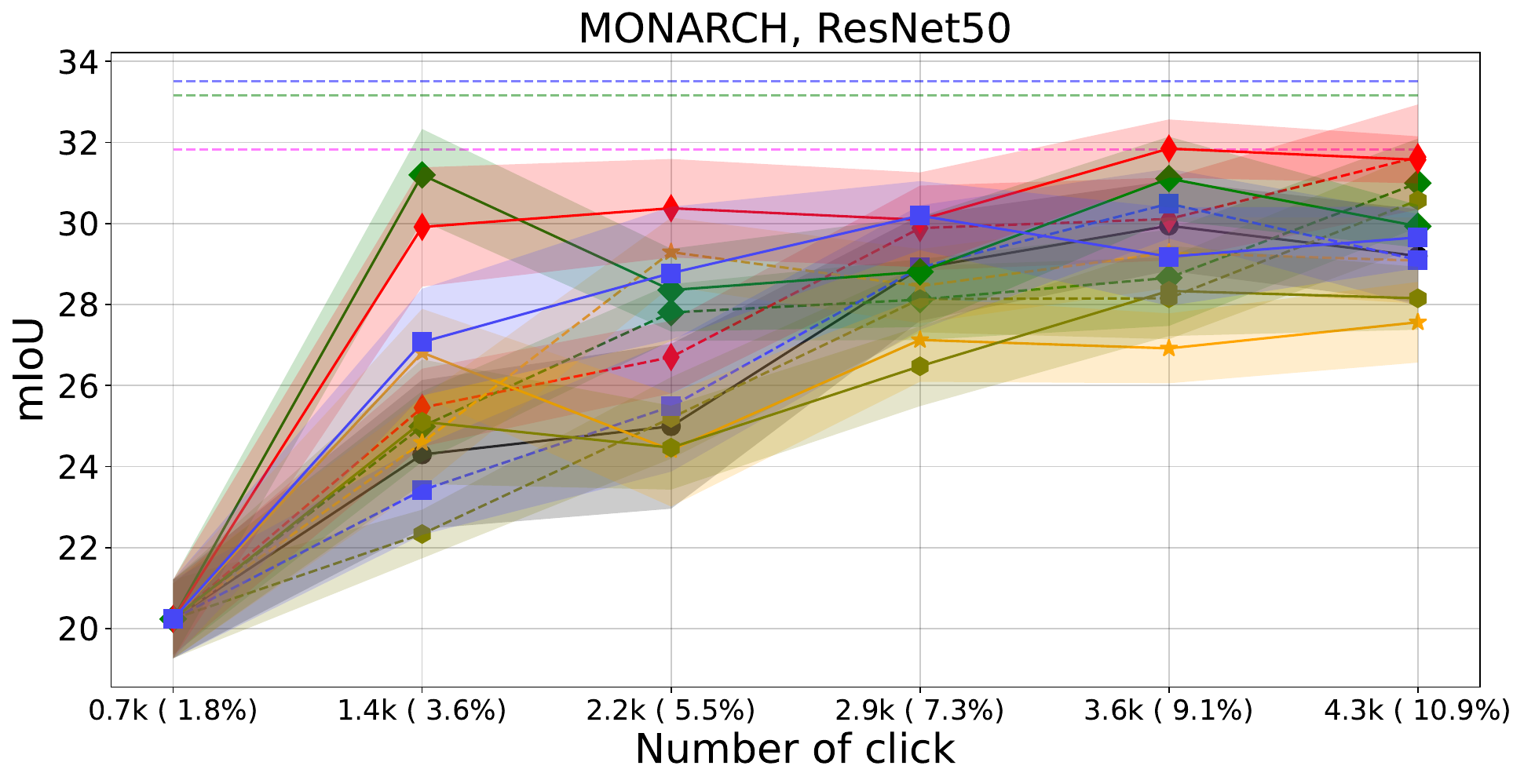}
    \includegraphics[width=0.5\linewidth]{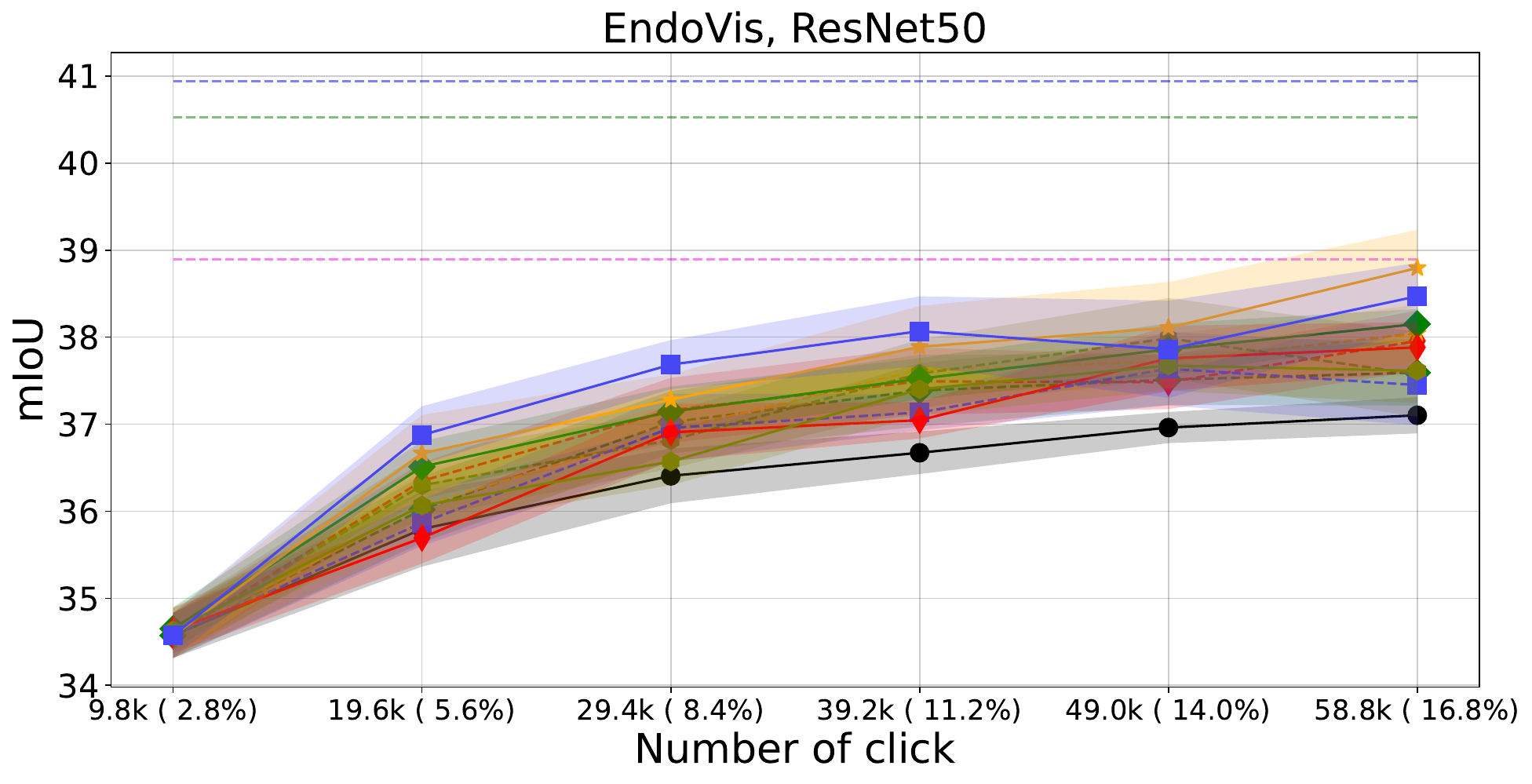}
   \caption{Comparison of different sampling strategies \RebuttalText{and their ablated version using the Max aggregation (Sec.~\ref{sec: max agg}}).
   Values are averaged over 3 training-validation splits for Cityscapes \cite{cityscapes}, Pascal VOC \cite{pascalVOC}, and 10 training-validation splits for EndoVis \cite{endovis}, MONARCH \cite{COWAL}. Error bars indicate one standard deviation. Results using ResNet101 and ViT backbones can be found in the Appendix.}
\label{fig: AL_results}
\end{figure*}

\begin{table*}[h!]
\centering
\begin{tabular}{l|c|c|c|c|c|c}
\hline
\textbf{Method} & \diagbox[width=8em]{\textbf{Backbone}}{\textbf{Datasets}} & \multicolumn{1}{c|}{\textbf{MONARCH}} & \multicolumn{1}{c|}{\textbf{PASCAL VOC}} & \multicolumn{1}{c|}{\textbf{ENDOVIS}} & \multicolumn{1}{c|}{\textbf{CITYSCAPES}} & \textbf{Average} \\ 
 & & mean - max & mean - max & mean - max & mean - max & mean - max \\
\hline
\hline
random & RN50 & 67.6 & 75.1 & 74.0 & 62.0 & 69.7 \\
BvSB \cite{BvSB}& RN50 & 67.2 - \TX{71.9} & 74.7 - \TX{77.0} & 74.9 - \TX{75.5} & \TX{65.4} - 65.1 & 70.6 - \TX{72.4} \\
revisiting SP \cite{revisiting}& RN50 & \TY{68.6} - \TY{\TX{73.7}} & \TY{75.9} - \TX{76.5} & \TY{\TX{75.2}} - 74.8 & 66.2 - \TX{66.7} & \TY{71.5} - \TY{\TX{72.9}} \\
Pixel Bal \cite{multi_label_AL}& RN50 & \TX{67.8} - 64.3 & 74.5 - \TX{77.0} & \TY{75.2} - \TX{76.0} & \TX{66.1} - 65.5 & \TX{70.9} - 70.7 \\
CBAL \cite{CBAL}& RN50 & \TX{64.3} - 64.0 & \TX{73.1} - \TX{73.1} & \TY{\TX{75.2}} - 74.8 & 66.9 - \TX{67.1} & \TX{69.9} - 69.8 \\
OREAL (ours) & RN50 & 66.2 - \TX{69.7} & 73.6 - \TY{\TX{77.2}} & 74.8 - \TY{\TX{76.2}} & \TY{67.9} - \TY{\TX{68.0}} & 70.6 - \TX{72.8} \\
\hline
random & RN101 & 74.5 & 75.4 & 72.2 & 59.7 & 70.5 \\
BvSB \cite{BvSB}& RN101 & 76.2 - \TY{\TX{77.2}} & 74.0 - \TX{77.3} & 73.1 - \TX{74.5} & 62.8 - \TX{63.3} & 71.5 - \TX{73.1} \\
revisiting SP \cite{revisiting}& RN101 & \TY{\TX{76.9}} - 75.5 & \TY{75.3} - \TX{77.0} & \TX{72.9} - \TX{72.9} & 63.7 - \TX{65.5} & 72.2 - \TX{72.7} \\
Pixel Bal \cite{multi_label_AL}& RN101 & \TX{75.7} - 75.2 & 73.6 - \TX{76.8} & 73.1 - \TY{\TX{75.0}} & 63.4 - \TX{65.5} & 71.5 - \TX{73.1} \\
CBAL \cite{CBAL}& RN101 & \TX{73.4} - 71.5 & \TX{73.6} - 72.3 & \TY{\TX{74.3}} - 74.0 & 65.5 - \TX{65.9} & \TX{71.7} - 70.9 \\
OREAL (ours) & RN101 & \TX{74.9} - 74.7 & 74.9 - \TY{\TX{77.9}} & 73.8 - \TX{74.8} & \TY{\TX{66.3}} - \TY{\TX{66.3}} & \TY{72.5} - \TY{\TX{73.4}} \\
\hline
random & ViT & 57.2 & 76.2 & 75.8 & 65.7 & 68.7 \\
BvSB \cite{BvSB}& ViT & 66.4 - \TX{72.1} & 72.4 - \TY{\TX{76.8}} & 76.2 - \TY{\TX{78.0}} & \TX{69.0} - 68.7 & 71.0 - \TY{\TX{73.9}} \\
revisiting SP \cite{revisiting}& ViT & 68.0 - \TX{72.0} & 72.1 - \TX{76.3} & \TY{76.6} - \TX{76.9} & \TX{69.5} - \TX{69.5} & 71.6 - \TX{73.7} \\
Pixel Bal \cite{multi_label_AL}& ViT & 67.3 - \TY{\TX{72.4}} & 71.8 - \TX{76.1} & 76.1 - \TX{77.9} & 69.6 - \TX{69.7} & 71.2 - \TX{73.8} \\
CBAL \cite{CBAL}& ViT & \TX{72.5} - 70.9 & 71.8 - \TX{72.7} & 76.2 - \TX{76.4} & \TY{\TX{71.1}} - \TY{70.1} & \TX{72.9} - 72.5 \\
OREAL (ours) & ViT & \TY{\TX{74.2}} - \TY{72.4} & \TY{73.2} - \TX{76.5} & 76.3 - \TX{76.9} & 68.5 - \TX{69.6} & \TY{73.1} - \TY{\TX{73.9}} \\
\hline
\hline
\textbf{Average} & (Excl. random) & 70.6 - \TX{71.8} & 73.6 - \TX{76.0} & 74.9 - \TX{75.6} & 66.8 - \TX{67.1} & 71.5 - \TX{72.7} \\
\hline
\end{tabular}
\caption{AuALC of all strategies across all datasets averaged over 10 runs
(MONARCH, EndoVis) and 3 runs (Cityscapes, Pascal VOC). Metrics are computed at the end of the 6 active learning steps on the respective test sets. For each strategy, the best score between using \MeanAgg~and \MaxAgg~is \RebuttalText{underlined}. For each backbone \RebuttalText{and each mean/max column}, the best method is \RebuttalText{in bold}.}
\label{tab: tab_results}
\end{table*}

\subsection{Implementation details}

\indent\indent
\textbf{Superpixels.} We use the SEEDS algorithm~\cite{SEEDS} to compute the superpixels of the training images. This computation is done once per dataset. For labels, we adopt the dominant labeling scheme from~\cite{revisiting}, where each superpixel is assigned the most prevalent label within its pixels allowing us to annotate a superpixel with a single click. According to \cite{revisiting}, this labeling approach is more efficient in terms of annotation cost compared to conventional annotation procedures for reaching higher model performance. EndoVis images are partitioned into 196~superpixels with an average size of $16^2$~pixels. For all the other datasets, superpixels have an average size of~$32^2$~pixels following optimal values found by previous work \cite{revisiting}. MONARCH images are partitioned into 36~superpixels, Cityscapes images into 576~superpixels, and Pascal VOC images into 256~superpixels.

\textbf{AL budget.} At each AL~iteration, our strategy selects~$50\times P$~superpixels, where $P$~is the number of superpixels per image. For MONARCH, we reduce the budget to $Q=20\times P$ to account for the faster saturation rate of its performance.

\textbf{Validation split}. For MONARCH and EndoVis, we randomly split the training videos into training and validation sets with a proportion of~$4:1$ for EndoVis, and~$2:1$ for MONARCH. For Cityscapes and Pascal VOC, we split the training images with a proportion of~$5:1$.

\textbf{Segmentation Model.} We use for the segmentation model~$f$, a DeepLabV3~\cite{deeplabV3} from the official PyTorch/vision GitHub repository~\cite{deeplab_pytorch} using a ResNet-50 and ResNet-101 backbone, and a vision transformer adapted from the work of \cite{Segmenter}.

\textbf{Training.} We adopt the training settings from~\cite{COWAL} for all datasets. We use a batch size of~4, a learning rate of~$10^{-4}$, and no weight decay using the Adam optimizer~\cite{adam}. At each step~$t$, the segmentation model~$f$ is trained with~$\A_t$ until the mIoU~score on the validation set converges, with a patience of~10. Validation is performed after each epoch, where an epoch is defined as the number of iterations required to cover the size of the whole training set, not just the size of~$\A_t$. We use Polyak averaging with~$\alpha=0.99$ for stable training evolution. For weight initialization, we use the ImageNet pre-trained weights when~$t=1$ and the weights obtained at the end of the step~$t-1$ when~$t>1$.

\textbf{Data augmentation}. For training our segmentation model, MONARCH images are resized to $220\times220$ and EndoVis images to $224\times224$, as specified in~\cite{COWAL}. Cityscapes images are resized to $769\times769$ and Pascal VOC images to $513\times513$, following the protocol in~\cite{revisiting}. After various augmentations detailed in the Appendix, all images are re-scaled back to their initial sizes.

\begin{figure*}[t]
    \includegraphics[width=\linewidth]{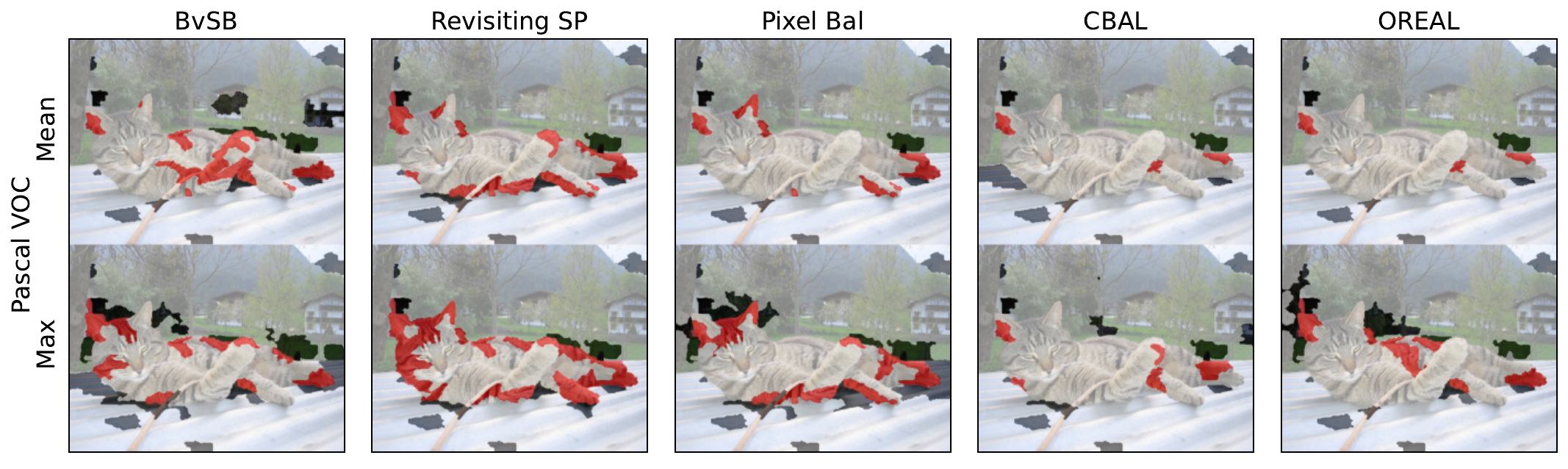}
   \caption{Selected superpixel regions of different sampling strategies. On average, we observe that when using the max aggregation, all sampling methods tend to select more regions around the boundary of objects. Additional plots can be found in the appendix.}
\label{fig:sampled patches}
\end{figure*}

\subsection{Evaluation protocol \& Results}
\label{sec: AuALC}

We perform 10~runs for each baseline on the MONARCH and EndoVis datasets, and 3~runs on the Cityscapes and Pascal VOC datasets. In each run, we perform 6~AL~steps and report the mean and standard deviation of the mIoU scores across all runs for each AL step. Following~\cite{COWAL}, we also report the area under the AL curve~(AuALC) as our evaluation metric. The AuALC is expressed as a fraction of the maximum possible area that a hypothetically perfect AL method would achieve if it could reach the same performance as training the model with the complete training dataset at each AL step.

Overall, the average performances of all AL methods are close with \OurSamp{}, which achieves the top AuALC on the RN101 and ViT backbones. \RebuttalText{However, \OurSamp{} is outperformed by the other baselines on the MONARCH dataset. Unlike the other datasets, the structure in the MONARCH dataset is simpler. Indeed, the surgery tools to segment always appears in the same position in the image and each image does not display more than one class (see Fig.~\ref{fig:auris}). This aspect of MONARCH could explain the different ranking of AL methods compared to other datasets}. As for the performance gap between using \MeanAgg~or \MaxAgg, the latter clearly outperforms the former (Fig.~\ref{fig: AL_results}, Tab.~\ref{tab: tab_results}). For instance, on the Pascal VOC dataset, the average AuALC for all AL methods and all backbones is 73.6 using \MeanAgg~while it increases to 76.0 when using \MaxAgg. In terms of mIoU, it corresponds to an average improvement of 1.63 across all AL iterations. See Appendix for more details.

\NewText{Figure~\ref{fig:sampled patches} shows a qualitative difference between patches sampled by different AL methods using both mean (\MeanAgg) and max (\MaxAgg) aggregation to calculate the superpixel uncertainty score. 
In addition to selecting objects of interest, this variation has also selected most of the neighbor regions around those objects. Since pixels close to the boundary between objects are the most difficult to classify, selecting those pixels for annotation greatly improves the prediction accuracy of both classes. We refer to Sec.~\ref{sec: max agg} for explaining this behavior of \MaxAgg.}

\NewText{\textbf{In summary, the quantitative and qualitative results we obtained empirically validate our hypothesis that creating context around an object for segmentation tasks improves segmentation accuracy.}}

\section{Discussion}
\textbf{Annotation Cost to reach 95\% Accuracy}. To further demonstrate the effectiveness of our approach, we also measure the annotation effort needed for each method to reach an accuracy of 95\% on the Pascal VOC dataset. Table~\ref{tab: annotation_effort} shows \MaxAgg~usually achieves this accuracy much faster than \MeanAgg. \newline

\NewText{\textbf{Superpixel Dominant Labeling}. Following previous work \cite{revisiting}, we divided our training images into superpixel regions and assigned one label to each region: the most prominent one. While superpixels tend to group semantically similar pixels together, they inevitably make errors and group pixels that are not from the same semantic object. As such, if we apply one label to the whole superpixel, we would create wrongly labeled pixels. We evaluate in this section the impact of these wrong labels. A model $A$ is trained on a dataset with its original semantic segmentation mask and a model $B$ is trained with the same dataset but whose segmentation masks are annotated using superpixel dominant labels. The performance of $B$ is the maximum score that any of the sampling methods we presented can achieve and is the full training score displayed in Fig.~\ref{fig: AL_results}. We show the performance of $A$ and $B$ in Tab.~\ref{tab: fulldatasetPr}. Surprisingly, except for Cityscapes, the performance gap between both models is rather small. We argue it is because, unlike Cityscapes, the other three datasets display simpler scenes which allow superpixel generation to be more precise. As such, fewer wrong pixel labels are created when assigning a single label to a whole superpixel on these types of datasets. See the Appendix for superpixel annotated labels.} \newline

\begin{table}[t]
\centering 
\resizebox{0.96\columnwidth}{!}{ 
\begin{tabular}{c | c | c | c}
    Datasets & \multicolumn{1}{c|}{ResNet50} & \multicolumn{1}{c}{ResNet101} & \multicolumn{1}{c}{ViT} \\
    & $A$ / $B$ & $A$ / $B$ & $A$ / $B$\\
    \hline
    Cityscapes & (67.5, 62.0) & (69.2, 61.3) & (61.2, 56.8)\\
    Pascal VOC & (57.5, 57.5) & (54.4, 53.9) & (69.7, 69.5) \\
    MONARCH & (35.3, 33.5) & (35.8, 33.5) & (40.7, 39.1) \\
    EndoVis & (42.1, 40.9) & (43.7, 42.0) & (35.6, 36.2) \\
    \hline
\end{tabular}
}
\caption{Comparison between original segmentation masks ($A$) and segmentation masks obtained using the dominant labeling scheme ($B$). A model is trained in both cases and evaluated on the same test set. The resulting mIoU is displayed here.}
\label{tab: fulldatasetPr}
\end{table}


\begin{table}[t]
\begin{tabular}{c | c | c}
    Methods & \multicolumn{1}{c|}{ResNet50} & \multicolumn{1}{c}{ResNet101} \\
    & mean / max & mean / max \\
    \hline
    Random & 26.6\% & 19.3\% \\
    BvSB \cite{BvSB}& (25.0\%, 17.4\%) & (22.6\%, 16.0\%) \\
    Revisiting SP \cite{revisiting}& (20.5\%, 18.0\%) & (21.2\%, 15.6\%) \\
    Pixel Bal \cite{multi_label_AL}& (24.6\%, 16.4\%) & (23.5\%, 16.5\%) \\
    CBAL \cite{CBAL}& (28.0\%, 28.5\%) & (21.2\%, 25.5\%) \\
    OREAL (ours) & (26.7\%, 17.4\%) & (20.5\%, 12.3\%) \\
    \hline
\end{tabular}
\caption{Amount of superpixel needed to reach 95\% of full dataset performance for each method on the Pascal VOC dataset.}
\label{tab: annotation_effort}
\end{table}


\textbf{Superpixel Weak Labeling}. Recently,~\cite{multi_label_AL} introduced a novel approach to weakly annotate superpixels. Rather than assigning a single dominant label to an entire superpixel, they identified all classes present within the superpixel without specifying the corresponding pixels for these classes. To leverage this weak labeling of superpixels, they proposed two innovative loss functions. The first, termed "Merged Positive Loss", motivates the model to predict any present class. The second, "Prototypical Pixel Loss", applies cross-entropy loss to the pixel with the highest confidence for each class, referred to as "prototypical pixels". 

Following training the model on the weakly labeled dataset (stage 1), \cite{multi_label_AL} suggests generating pseudo labels using these prototypical pixels and they retrain the model by including these pseudo labels (stage 2). They compared the feature representation of each pixel to the prototypical pixels and assigned the label of the closest prototype in the feature space to the pixels. Their proposed sampling method, \PixelBalSamp, demonstrated state-of-the-art performance on the Cityscapes \cite{cityscapes} and Pascal VOC \cite{pascalVOC} datasets, as shown in their results. 

\NewText{In a new experiment, we replace the dominant labeling scheme with the weak labeling scheme and show results in Tab.~\ref{tab: MUL resutls}. \OurSamp~ outperformed \PixelBalSamp~on the Cityscapes dataset and remained competitive on Pascal VOC. The performance gap between using \MeanAgg~and \MaxAgg~is marginal on Pascal VOC but becomes significant on Cityscapes. These results further validate our hypothesis that creating context around objects improves the segmentation accuracy.}



\begin{table}[h!]
\centering
\begin{tabular}{l|c|c|c}
Method & Stage & \multicolumn{1}{c|}{Pascal VOC} & \multicolumn{1}{c}{Cityscapes} \\ 
 & & mean - max & mean - max \\
\hline
\hline
random & 1 & 73.5 & 71.1 \\
BvSB \cite{BvSB}& 1 & \TX{75.4} - 75.2 & 72.3 - \TX{72.8} \\
revisiting SP \cite{revisiting}& 1 & 75.6 - \TY{\TX{76.2}} & 72.7 - \TX{73.9} \\
Pixel Bal \cite{multi_label_AL}& 1 & \TX{75.6} - 75.4 & 73.4 - \TX{73.7} \\
OREAL & 1 & \TY{\TX{75.7}} - 75.5 & \TY{74.0} - \TY{\TX{74.4}} \\
\hline
random & 2 & 75.0 & 74.2 \\
BvSB \cite{BvSB}& 2 & \TX{76.6} - 76.5 & 74.9 - \TX{75.4} \\
revisiting SP \cite{revisiting}& 2 & \TX{76.5} - 76.2 & 75.1 - \TX{76.6} \\
Pixel Bal \cite{multi_label_AL}& 2 & \TY{\TX{76.7}} - 76.2 & 76.0 - \TX{76.5} \\
OREAL & 2 & 76.5 - \TY{\TX{76.6}} & \TY{76.6}. - \TY{\TX{76.8}} \\
\hline
\hline
\textbf{Average} & & \TX{76.1} - 76.0 & 74.4 - \TX{75.0} \\
\hline
\end{tabular}
\caption{AuALC of all strategies using the weak labeling scheme. For each strategy, the best score between using \MeanAgg~and \MaxAgg~is \RebuttalText{underlined}. For each stage, the best strategy is \RebuttalText{in bold}.}
\label{tab: MUL resutls}
\end{table}
\section{Conclusion}
\NewText{We propose an approach for achieving context sampling in the case of semantic segmentation tasks as well as OREAL, a novel active learning sampling strategy. We suggest it is important to sample the context around objects of interest since the most difficult pixels to classify are located at the boundaries between objects. We validated these hypothesis through extensive experiments.}

{\small
\bibliographystyle{ieee_fullname}
\bibliography{main}
}

\end{document}